\documentclass[letterpaper, 10pt, conference, twoside]{ieeeconf}    

\makeatletter
\let\NAT@parse\undefined
\makeatother

\usepackage{dblfloatfix}

\usepackage[numbers,sectionbib,sort&compress]{natbib}
\usepackage{bm}
\usepackage{gensymb}
\usepackage{xcolor}
\usepackage{graphicx}
\usepackage{amsmath}
\usepackage{amssymb}
\usepackage{amsfonts}
\usepackage{subcaption}
\usepackage{amsfonts}
\usepackage{siunitx}
\usepackage{booktabs}
\usepackage{makecell}
\usepackage{multirow}
\usepackage{upgreek}
\usepackage[font=small]{caption}
\usepackage[export]{adjustbox}
\usepackage{tikz}
\usepackage{tabularx}
\usepackage{svg}
\usepackage{float}
\usepackage{sidecap} \sidecaptionvpos{figure}{c}
\usepackage{hyperref}
\hypersetup{colorlinks,breaklinks,
linkcolor=[rgb]{0.5,0.,0.},
citecolor=[rgb]{0.000,0.427,0.173},
urlcolor=[rgb]{0.031,0.318,0.612}}

\usepackage[nameinlink, capitalize]{cleveref}
\usepackage{color}
\definecolor{CommentPink}{rgb}{1,0.2,0.5}
\definecolor{CommentBlue}{rgb}{0,0,1}
\definecolor{CommentGreen}{rgb}{0,1,0}

\Crefname{section}{Sec.}{Sec.}
\Crefname{equation}{Eq.}{Eq.}

\newcommand\blfootnote[1]{%
\begingroup 
\renewcommand\thefootnote{}\footnote{#1}%
\addtocounter{footnote}{-1}%
\endgroup 
}

\newcommand{\etal}{\textit{et al}.}
\newcommand{\ie}{\textit{i}.\textit{e}., }

\usepackage[printonlyused,withpage,nolist,nohyperlinks]{acronym}

\IEEEoverridecommandlockouts   

\title{\LARGE \bf Active Implicit Reconstruction Using One-Shot View Planning}

\author{Hao Hu$^{\star}$ \and Sicong Pan$^\star$ \and Liren Jin \and Marija Popović \and Maren Bennewitz %
}

\begin{document}
\maketitle
\thispagestyle{empty} 
\pagestyle{empty}

\begin{abstract} 
Active object reconstruction using autonomous robots is gaining great interest. A primary goal in this task is to maximize the information of the object to be reconstructed, given limited on-board resources. Previous view planning methods exhibit inefficiency since they rely on an iterative paradigm based on explicit representations, consisting of (1) planning a path to the next-best view only; and (2) requiring a considerable number of less-gain views in terms of surface coverage. To address these limitations, we propose to integrate implicit representations into the One-Shot View Planning (OSVP). The key idea behind our approach is to use implicit representations to obtain the small missing surface areas instead of observing them with extra views. Therefore, we design a deep neural network, named OSVP, to directly predict a set of views given a dense point cloud refined from an initial sparse observation. To train our OSVP network, we generate supervision labels using dense point clouds refined by implicit representations and set covering optimization problems. Simulated experiments show that our method achieves sufficient reconstruction quality, outperforming several baselines under limited view and movement budgets. We further demonstrate the applicability of our approach in a real-world object reconstruction scenario.
\vspace{-0.5cm}
\end{abstract} 

\blfootnote{$^\star$These authors contributed equally to this work.}%
\blfootnote{Hao Hu is with Intel Asia-Pacific Research \& Development Ltd. Sicong Pan and Maren Bennewitz are with the Humanoid Robots Lab, Liren Jin and Marija Popović are with the Institute of Geodesy and Geoinformation, University of Bonn, Germany. Maren Bennewitz is additionally with the Lamarr
 Institute for Machine Learning and Artificial Intelligence,
 Germany. This work has partially been funded by the Deutsche Forschungsgemeinschaft (DFG, German Research Foundation) under BE 4420/4-1 within the FOR 5351 – 459376902 – AID4Crops and under Germany’s Excellence Strategy, EXC-2070 – 390732324 – PhenoRob. Corresponding: \href{mailto:span@uni-bonn.de}{span@uni-bonn.de}
}%


\section{Introduction} \label{secI}

Active object reconstruction~\cite{chen2011active} involves a robot system actively placing sensor views to acquire the information about an object to be reconstructed. One primary goal is to maximize reconstruction quality while respecting limited on-board resources, such as movement cost and view budget~\cite{song2021view}. The view planning problem in such tasks is often solved by an iterative paradigm~\cite{zeng2020view}, where the most informative next-best-view (NBV) is selected iteratively based on the current state of an explicit representation~\cite{delmerico2018comparison, pan2021global}. While iterative paradigms can yield high-quality reconstructions, specifically by achieving high surface coverage, they tend to be inefficient since they rely on frequent 3D representation updates and iteratively planning a path to the NBV.

In the recent work SCVP~\cite{pan2022scvp}, a one-shot pipeline is proposed to avoid iterative view planning, inspired by the insight of the set covering optimization problem (SCOP)~\cite{kaba2017reinforcement}, which finds the smallest set of views satisfying the full coverage of object surfaces. Specifically, a neural network is utilized to predict the ideally smallest set of views given the occupancy voxel map generated by initial point cloud measurements. Since all required views are planned in the predicted set of views, a globally shortest path can be computed to overcome the inefficiencies of iterative planning.

\begin{figure}[!t]
\centering
\includegraphics[width=1.0\columnwidth]{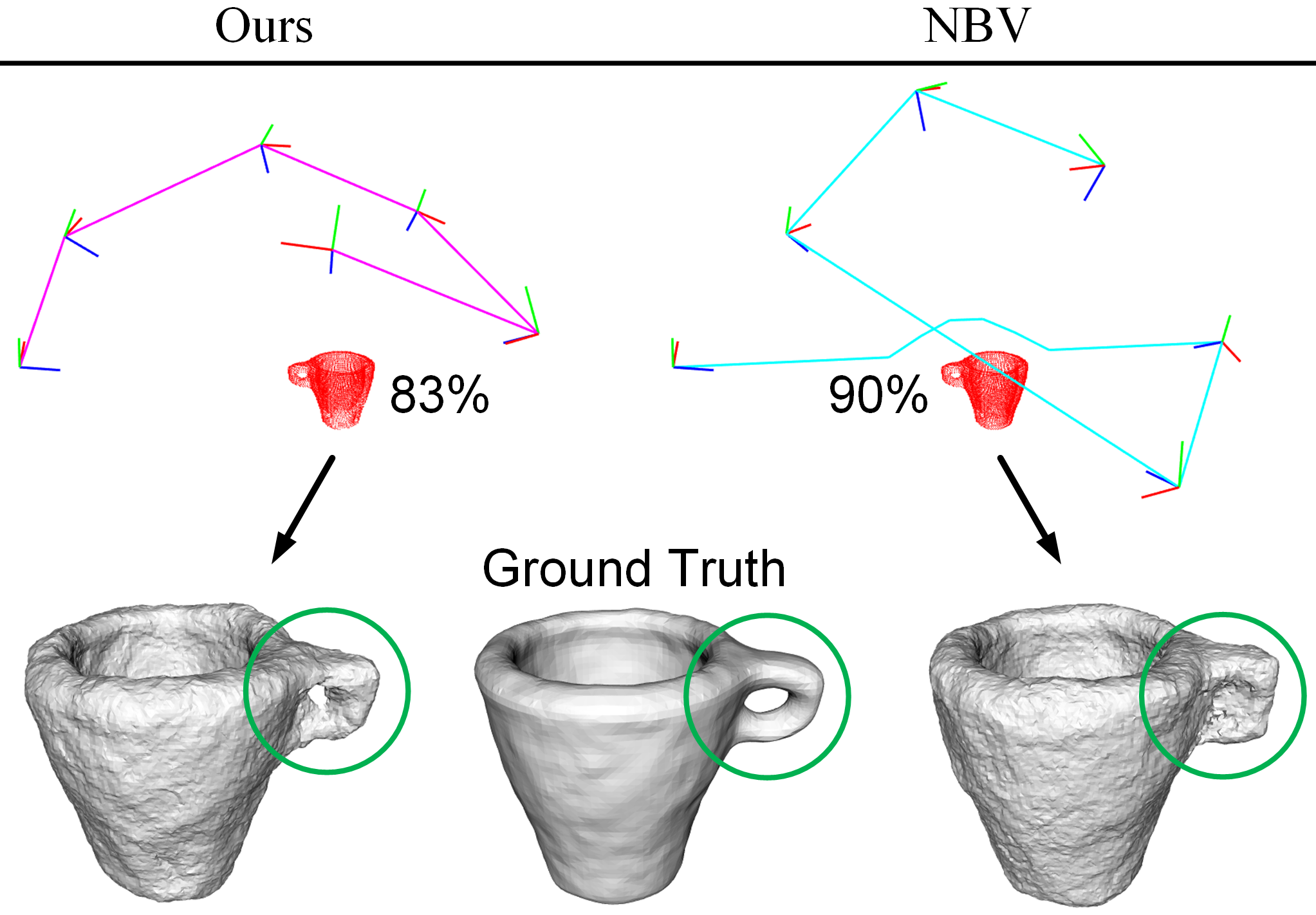}
\caption{
A comparison of our proposed implicit representation-based one-shot view planning and an explicit representation-based next-best-view (NBV) planning method~\cite{pan2022aglobal} under a limited view budget (six in this example). The views (red-green-blue), global path (purple), local path (cyan), and accumulated observed point clouds along with their surface coverage ($83\%$ and $90\%$) are shown on the top regions. The bottom row shows reconstructed meshes using the implicit representation~\cite{Boulch_2022_CVPR}. Compared to the ground truth mesh, our method achieves a comparable mesh quality, especially in the handlebar area of the cup (green circles), with less surface coverage and less movement cost than the NBV baseline.
}
\label{fig_paper_cover}
\vspace{-0.5cm}
\end{figure}

In most view planning approaches, an explicit map representation is required as the basis for planning. However, in explicit representations, only the object surfaces observed by sensors are considered reconstructed, \ie what you measure is what you get. Since certain complex surface areas of an object often require specific viewpoints to observe, the gain of surface coverage within the view sequence follows a long-tailed distribution~\cite{pan2023one}. In other words, most surface areas of the object can be effectively covered with just a few views. Achieving a good explicit 3D representation usually requires a high surface coverage. As a result, a considerable number of less-gain views in terms of surface coverage is planned in existing works~\cite{delmerico2018comparison,mendoza2020supervised,zeng2020pc,pan2022aglobal,pan2022scvp,pan2023global,pan2023one}.

To alleviate this problem, we propose utilizing implicit neural representations, which use deep neural networks to learn a continuous surface function. By training on large 3D object datasets, these networks can learn prior knowledge about geometric primitives. This endows implicit neural representations such as occupancy networks~\cite{peng2020convolutional, tang2021sa, Boulch_2022_CVPR} with the capability to interpolate and extrapolate on 3D surfaces. Thus the small areas observed by the above-mentioned less-gain views can be replaced by sampling from a learned continuous surface function. To this end, we employ the state-of-the-art point cloud-based implicit surface reconstruction approach, POCO~\cite{Boulch_2022_CVPR}, on the RGB-D data as our system inputs.

The primary focus of this paper is to bridge the gap between implicit representations and the highly efficient one-shot view planning pipeline. While SCVP~\cite{pan2022scvp} addresses one-shot planning, it is based on an explicit voxel-based set covering optimization, which is not applicable within the context of implicit representations. The challenge stems from the fact that the quality of implicit representations and surface coverage do not exhibit a strictly positive correlation. A sparse point cloud observation with low surface coverage can still yield good implicit reconstructions. Therefore, the distribution of viewpoints outweighs the pure surface-coverage optimization objective for reconstructing an implicit representation. 

Based on this insight, we propose a new approach to generate a training dataset to learn one-shot view planning. Instead of relying on constraints related to full surface coverage based on observations, we shift our optimization objective to finding the ideally smallest set of views to achieve full coverage based on refined dense surface points using implicit representations. We introduce a novel network, named OSVP, to learn the prediction. After obtaining an initial sparse point cloud, OSVP takes a dense point cloud refined by POCO~\cite{Boulch_2022_CVPR} as input and predicts the ideally smallest set of views. These views are then connected using global path planning to reduce the movement cost.

Fig.~\ref{fig_paper_cover} provides an illustration of our approach. Compared to a traditional NBV-based method, our method traverses a shorter global path under the same view budget and achieves a comparable quality of implicit representation even with a lower surface coverage. We perform experiments compared to several state-of-the-art NBV-based~\cite{delmerico2018comparison,mendoza2020supervised,zeng2020pc,pan2022aglobal,pan2023global} and one-shot view planning methods~\cite{pan2022scvp,pan2023one}. To analyze the reconstruction efficiency, we highlight the results with limited budgets of resources in terms of both the required number of views and movement cost. The contributions of our work are threefold:
\begin{itemize}
\item An efficient one-shot view planning method for active implicit reconstruction, achieving sufficient reconstruction quality with limited views and movement budgets. 
\item A new dataset generation approach for labeling the smallest set of views, covering dense surface points refined from implicit representations.
\item Our OSVP network predicts a small set of views at once by learning features of dense surface points refined from implicit representations.
\end{itemize}
To support reproducibility, our implementation and dataset is published at \url{https://github.com/psc0628/AIR-OSVP}.

\section{Related Work} \label{secII}

Robotic active reconstruction usually incorporates two major components: view planning and 3D representation. In this section, we provide a brief overview of relevant works using different planning methods and 3D representations for active reconstruction tasks.


\subsection{View Planning in Active Reconstruction} \label{secII.B}
For reconstructing an unknown object, a common approach is to iteratively plan the NBV to maximize the expected utility~\cite{potthast2014probabilistic, daudelin2017adaptable, zaenker2021viewpoint, menon2022viewpoint, zaenker2023graph}. Delmerico \etal~\cite{delmerico2018comparison} evaluate the utility of candidate views by calculating the probability of observing objects within its field of view. Pan \etal~\cite{pan2022aglobal,pan2023global} improve the utility calculation by defining global surface coverage optimization. 
Without relying on hand-crafted utility calculation, deep learning-based approaches learn the utility function by training on datasets. Mendoza \etal~\cite{mendoza2020supervised} use a 3D
convolutional neural network, called NBVNet, to score view candidates, given input occupancy voxel map. PCNBV~\cite{zeng2020pc} takes point cloud and the current view selection states as input, subsequently estimating the utility for all candidate views. Other methods use reinforcement learning~\cite{peralta2020next,zeng2022deep,dengler2023viewpoint} to learn a policy for view planning.

SCVP~\cite{pan2022scvp} is the first implementation of one-shot view planning for unknown object reconstruction. Given the initial point cloud observation, a trained network predicts the global view configuration at once. In follow-up work, a novel multiview-activated architecture MA-SCVP~\cite{pan2023one} is introduced along with an efficient dataset sampling method for view planning based on a long-tail distribution to further improve the reconstruction efficiency. In this work, we follow the same idea to predict a set of views for planning global paths in a one-shot paradigm. However, a key difference is that we find the smallest set of views to achieve full coverage based on refined dense surface points using implicit representations rather than observed sparse points, which makes our reconstruction more efficient and ensures the quality of the reconstructed mesh.

\begin{figure*}[!t]
\centering
\includegraphics[width=1.0\textwidth]{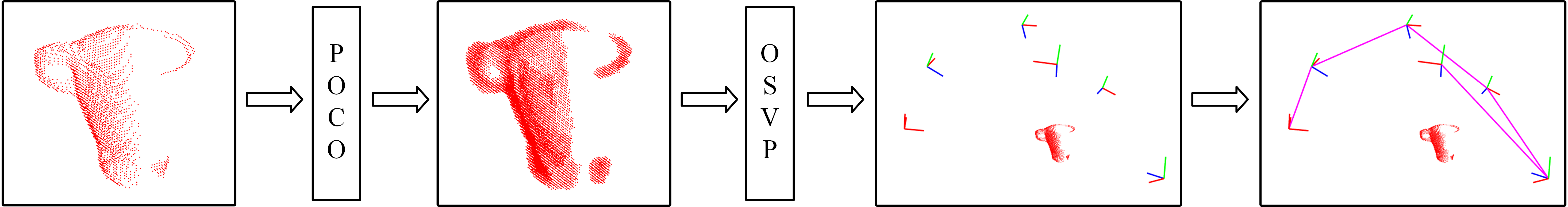}
\caption{An example of our online workflow. The robot observes a sparse point cloud at the initial view, which is passed through POCO to generate a refined point cloud. We input the refined point cloud into our OSVP network to predict a set of views for efficiently covering the object. The planned views (red-green-blue) and the initial view (red) are connected by a global path (purple) for the robot to execute.}
\label{fig:pipeline}
\vspace{-0.5cm}
\end{figure*}

\subsection{3D Representations for Active Reconstruction} \label{secII.A}

Previous 3D object reconstruction methods generally use explicit representations, including data structures such as voxel grids~\cite{delmerico2018comparison}, point clouds~\cite{zeng2020pc}, and meshes~\cite{wu2014quality}. These explicit representations usually discretize 3D space, leading to significant computational cost and memory burden to obtain enough surface details, especially for large-scale reconstruction and environments with high-fidelity details.

Recently, implicit neural representations have gained significant interest in object reconstruction, due to their low memory consumption and strong representation ability.

They can be realized by a continuously differentiable function in the form of neural networks. Mildenhall \etal~\cite{mildenhall2021nerf} use only posed images to learn a neural radiance field (NeRF) for view synthesis. DeepSDF~\cite{park2019deepsdf} represents a shape's surface by a continuous volumetric field implemented by a deep neural network, in which the predicted magnitude of a point represents the distance to the surface boundary. Peng \etal~\cite{peng2020convolutional} propose a convolutional neural network to predict the occupancy probability of any point in 3D space. Similarly, POCO~\cite{Boulch_2022_CVPR} performs information aggregation by applying convolution on latent feature vectors saved at each point position in input point clouds. It then utilizes a convolution-based interpolation on the nearest neighbors, employing inferred weights to enhance the provided point cloud, resulting in a more complete and detailed implicit representation. In this work, we leverage this generalizable convolutional approach for interpolating and extrapolating on 3D surfaces to alleviate the aforementioned long-tailed problem~\cite{pan2023one}.

Integrating implicit representations into online active view planning frameworks is an open challenge. Early attempts in this direction apply NBV planning in implicit representations~\cite{yan2023active, sunderhauf2023density, ran2023neurar}, following a greedy approach. However, using NBV with such representations is inherently limited in computational and planning efficiency since network retraining is usually necessary between planning steps. To address this problem, we propose merging one-shot planning paradigms with implicit representations. This enables us to accelerate planning performance while maintaining high implicit reconstruction quality. 


\section{System Overview} \label{secIII}

Our goal is to reconstruct an implicit representation of an unknown object located on a tabletop. This reconstruction is accomplished by utilizing a series of point clouds captured from planned sensor views in a discrete hemispherical candidate view space $V=\{v_i\,|\,i=1,2,\ldots,n\}\subset\mathbb{R}^3\times SO(3)$, where $n$ is the number of view candidates. The position of each view is defined by the Tammes problem~\cite{lai2023iterated}, which finds the placement of a given number of points on a sphere to maximize the minimum distance between them. The pose of a candidate view is viewing the center of the object as usually considered in active object reconstruction approaches~\cite{mendoza2020supervised,zeng2020pc,pan2022scvp}.

Fig.~\ref{fig:pipeline} shows the workflow of the online phase in our object reconstruction pipeline. The online phase begins with the robotic acquisition of a point cloud from an initial view $v_{init} \in V$, which can be randomly selected or defined by a user. We refine the sparse initial point cloud observation by POCO~\cite{Boulch_2022_CVPR} to obtain a dense point cloud with richer surface details. The benefit of using POCO as a refinement module is studied in Sec.~\ref{secV.C}. Given this refined point cloud, our OSVP network predicts a set of views for data acquisition. Finally, a global path is calculated for traversing these planned views starting from the initial view. The robot then navigates to each view along the global path to capture point clouds. In the offline phase, these point clouds and view poses are used for the POCO-based implicit reconstruction.

Our global path planning method solves the problem of connecting all planned views starting from the initial view. We compute the optimal global path by finding the shortest Hamiltonian path on a graph. We use state compression dynamic programming \cite{held1962dynamic} to solve this problem within seconds since the number of views is usually small. An illustration of the global path is given in Fig.~\ref{fig:pipeline}. The view-to-view local path follows the concept of avoiding the object as an obstacle on the tabletop as fully defined in~\cite{pan2023one}.

\section{Learning One-Shot View Planning via Implicit Surface} \label{secIV}

This section presents our new OSVP network, which is designed to adaptively predict a set of views based on a dense point cloud refined by POCO. For network training, we generate a dataset consisting of the partially observed object point cloud and its ground truth view set. 

\subsection{Refining Surface Points via POCO} \label{secIV.A}

\begin{figure}[!t]
\centering
\includegraphics[width=0.55\columnwidth]{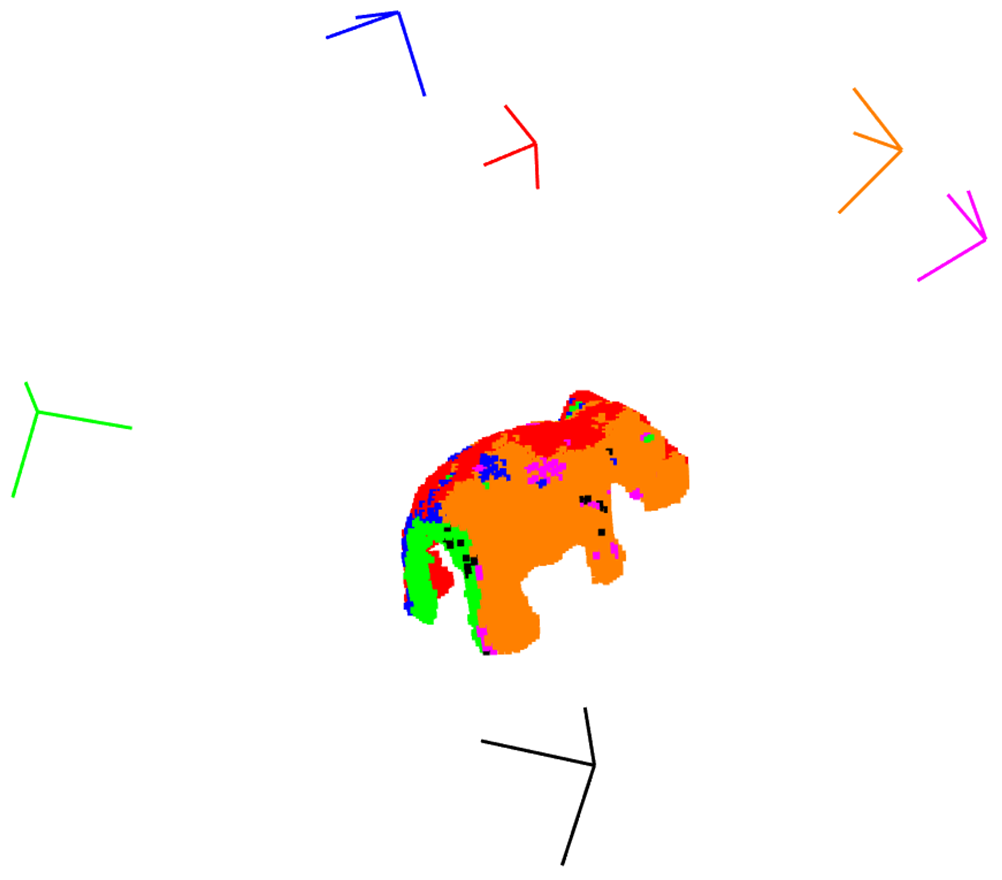}
\caption{
An example solution to SCOP for object surface coverage using six views. Colors represent views and the corresponding surfaces of the object they cover.
} 
\label{fig_set_covering}
\vspace{-0.5cm}
\end{figure}

Implicit representations are known for favoring a continuous representation of an object's surface. In our approach, we use POCO~\cite{Boulch_2022_CVPR} to generate an implicit representation given initial point cloud measurements. POCO takes in a set of 3D points sampled on a surface and tries to construct a continuous function $f:\mathbb{R}^3\rightarrow[0, 1]$ indicating the occupancy of a given 3D query point. POCO is trained on the ShapeNet dataset~\cite{shapenet2015} to learn the prior knowledge of geometric primitives. Leveraging this feature, we can refine sparse point cloud observations by exploiting POCO's learned prior knowledge to get a dense object surface with more details as illustrated in Fig.~\ref{fig:pipeline}.

These dense surface points are obtained by densely querying the occupancy prediction from the pre-trained POCO network. Subsequently, they are voxelized using OctoMap~\cite{hornung2013octomap}, where its ray-casting plays a crucial role in addressing the subsequent SCOP. By transferring the continuous occupancy field of POCO into explicit point clouds, we can now refer to SCOP~\cite{kaba2017reinforcement} for covering the object surfaces. Furthermore, the number of views required for full surface coverage can be reduced by these dense point clouds, since more surface detail information is already included in each view. 

\subsection{Covering Object Surface via Refined Points} \label{secIV.B}

Assuming there is a ground truth mesh from 3D object datasets, we first sample the mesh into dense surface points, denoted as $P_{gt}$. Next, the point cloud observed from each view $v_i$ is refined by POCO, denoted as $P_{i}$. We fuse $P_i$ to obtain the full set of point cloud $P_{\mathit{full}}$ for set covering. Since there are some noisy points in POCO-refined surfaces, we include a hyper-parameter $\alpha\in\mathbb{N}^{+}$ to implement the outlier removal mechanism, making a reasonable set covering. Whenever a 3D point $p$ is present in more than $\alpha$ views, we consider it to be essential for constructing the object surface:
\begin{equation}
\label{eq:equ1}
\begin{aligned}
P_{\mathit{full}}=\{p \mid p \in (\bigcup_{i=1}^{|V|} P_i \cap P_{gt}) \land \mathrm{count}(p) \geq \alpha \},
\end{aligned}
\end{equation}
where $V$ is the candidate view space and the function $\mathrm{count}$ calculates how many times the point $p\in P_{\mathit{full}}$ appears in each point cloud $P_i$.
Subsequently, given refined points $P_i$, we can compute the smallest set of views that covers $P_{\mathit{full}}$ by formulating a SCOP, which is then solved by Gurobi~\cite{gurobi2021gurobi}:

\begin{equation}
\label{equ2}
\begin{aligned}
\mathrm{minimize}:& \sum_{v\in V} z_v \\
\mathrm{subject\ to}:&\ (a)\ \sum_{p\in{P_i}} z_{v}\geq1\ \mathit{for\ all}\ p\in P_{\mathit{full}} \\
            &\ (b)\ z_{v}\in\{0,1\}\ \ \mathit{for\ all}\ {v}\in {V}
\end{aligned}
\end{equation}
The objective function $\sum_{v\in V} z_v$ minimizes the number of chosen views. It is subject to (a) each point element $p\in P_{\mathit{full}}$ must be covered by at least one chosen view that belongs to this point $p\in {P_i}$, and (b) $z_{v}$ is a binary variable indicating that each view is either in the set of chosen views or not. We illustrate an example of the smallest set of views that fully cover all object surfaces in Fig.~\ref{fig_set_covering}.

\begin{figure}[t]
\centering
\includegraphics[width=1.0\columnwidth]{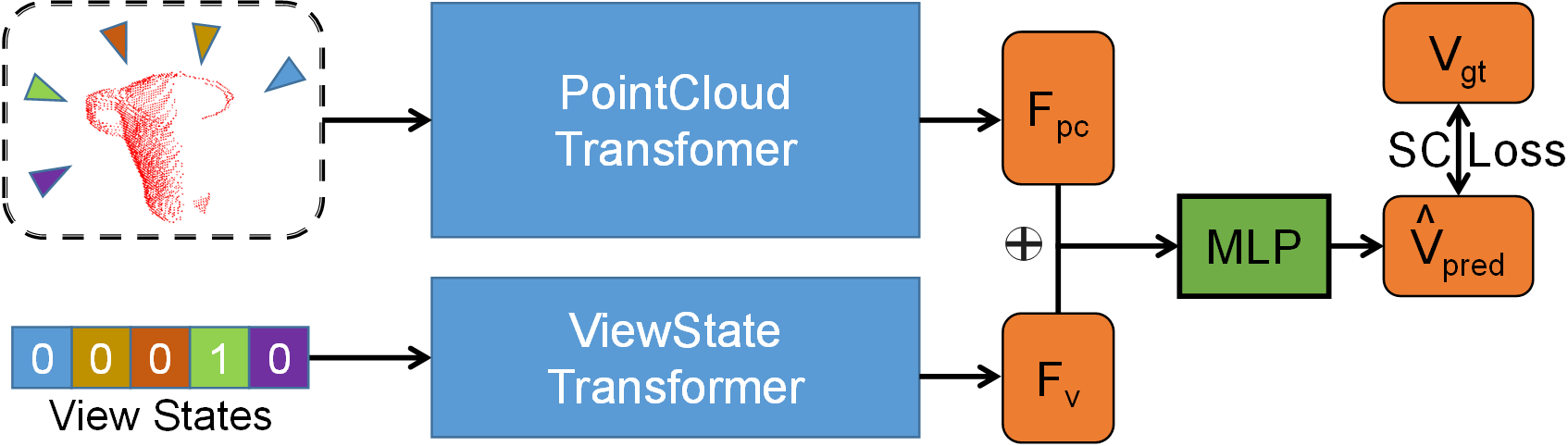}
\caption{OSVP network architecture: PoinTr~\cite{yu2021pointr} is used as the backbone to extract features from a point cloud and a vanilla 1-D transformer~\cite{vaswani2017attention} is used to process view states. $\bigoplus$ denotes element-wise addition. MLP stands for fully connected layers. The SCLoss is computed between $\hat{V}_{pred}$ and $V_{gt}$ for training.
}
\vspace{-0.5cm}
\label{fig:net_arch}
\end{figure}

\subsection{Generating Set-Covering Dataset} \label{secIV.C}

For an object surface points $P_i$ as input, the goal of OSVP network is to predict the smallest set of views to cover the remaining surface, \ie $P_{\mathit{full}}\setminus P_i$. For example, if the input is the red view in Fig.~\ref{fig_set_covering}, the smallest set of views then becomes the set of five remaining views. We follow the long-tailed sampling method~\cite{pan2023one} to generate better input cases of object surface points and solve SCOP for them.

\subsection{Network Architecture and Loss Function} \label{secIV.D}

Our OSVP network predicts the smallest set of views for reconstruction from a discretized view space. Fig.~\ref{fig:net_arch} shows the architecture of the OSVP network. We employ the PointCloud Transformer module from the PoinTr~\cite{yu2021pointr} network to extract features from the input point cloud. As demonstrated in PCNBV~\cite{zeng2020pc} and MA-SCVP~\cite{pan2023one}, incorporating the view state improves the network prediction. Therefore, we incorporate a ViewState Transformer module to process the input view states. The view state is a binary mask that indicates whether a particular view is utilized in the observation of the initial point cloud or not. These encoded vectors from both transformers are then summed and further processed by fully connected layers to predict the probability of each view being chosen or not. 


As discussed in SCVP~\cite{pan2022scvp}, the 1's and 0's in the ground truth mask $V_{gt}$, which is a binary mask of the smallest set solved by SCOP, have different importance to the network training. We use a hyper-parameter $\lambda$ to balance them. Our set-covering SCLoss can be described as:

\begin{equation}
\label{equ3}
\begin{footnotesize}
\begin{aligned}
\mathit{CE_i} &= v_i^{gt}logv_i^{net}+(1-v_i^{gt}) log(1-v_i^{net}), \\
\mathit{SCLoss} &= -\frac{1}{|{V}|}\sum_{i}^{|{V}|}[(1-v_i^{gt})\times \mathit{CE_i} + \lambda \times v_i^{gt}\times \mathit{CE_i}],
\end{aligned}
\end{footnotesize}
\end{equation}
where $\mathit{CE_i}$ is cross entropy loss term from $v_i$ prediction, $v_i^{gt} \in V_{gt}$ and $v_i^{net}$ is the element from the network prediction $\hat{V}_{pred}$, which denotes the probability of choosing $v_i$ or not. If $v_i^{net}$ exceeds a value of 0.5, we consider this view as necessary for the following reconstruction. This value of 0.5 is experimentally confirmed in SCVP~\cite{pan2022scvp} to achieve a good network performance.


\section{Experimental Evaluation}\label{secV}

\begin{figure}[!t]
\centering
\includegraphics[width=1.0\columnwidth]{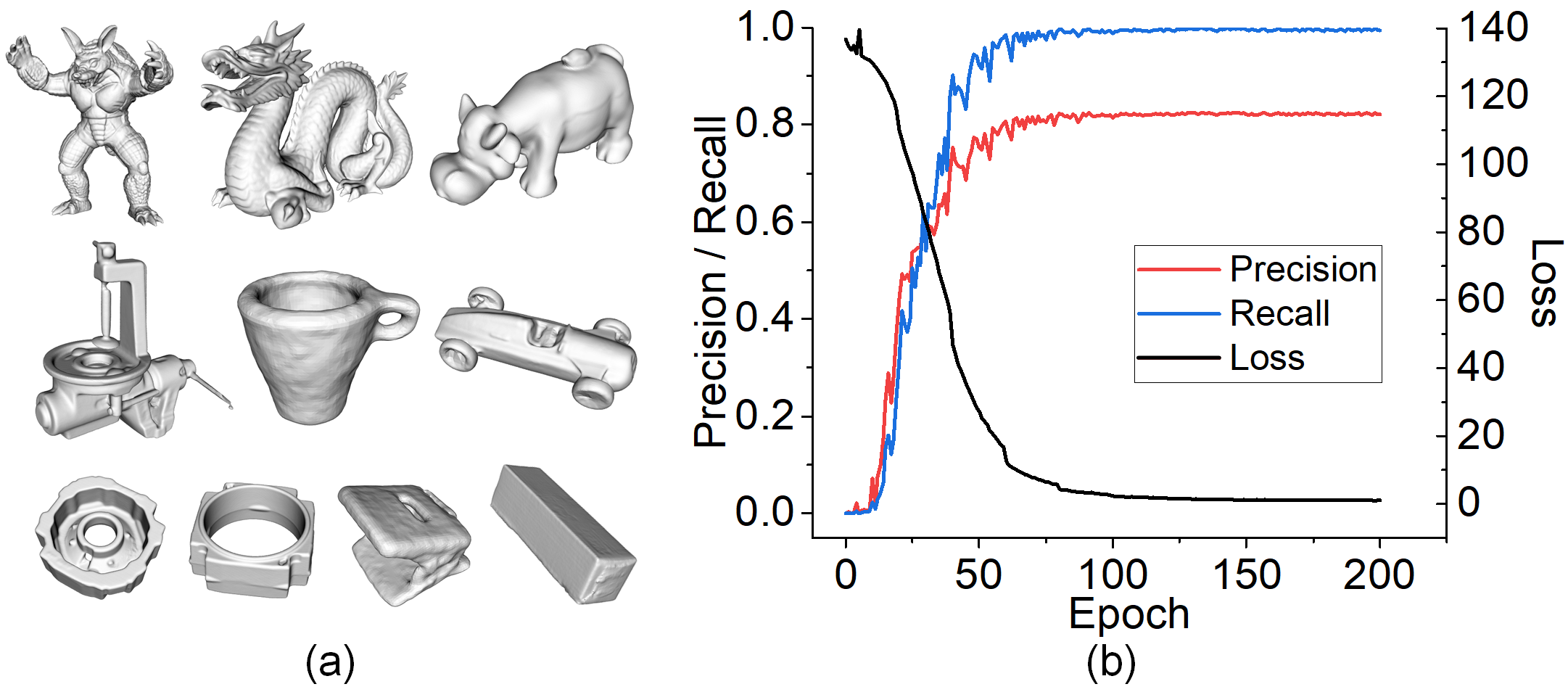}
\caption{Test objects and OSVP network training: (a) 3D mesh models of 10 test objects with complex surfaces; (b) precision, recall, and loss on the validation dataset over training epochs.
}
\label{evp_metrics}
\vspace{-0.5cm}
\end{figure}

Our experiments are designed to support the claim that our method can achieve a good implicit reconstruction of an unknown object within a limited budget of resources by one-shot view planning. The experiments are conducted on an Intel Core i7 12700K CPU, 32GB of RAM, and a GeForce RTX 3070Ti graphics card with 8GB memory. 

\begin{table}[!t]
\centering
\resizebox{0.85\columnwidth}{!}{%
\begin{tabular}{|c|c|c|c|c|c|}
\hline
$\lambda$    & 0.75  & 1.00  & \textbf{1.25}  & 1.50  & 2.00  \\ \hline
Recall    & 67.09 & 87.86 & 99.31 & \textbf{99.80} & 98.70 \\ \hline
Precision & \textbf{97.80} & 91.96 & 82.35 & 59.78 & 60.43 \\ \hline
F1 Score  & 79.59 & 89.86 & \textbf{90.04} & 74.77 & 74.96 \\ \hline
\end{tabular}
}
\caption{Ablation study of hyper-parameter $\lambda$ on the validation dataset. We choose the network with $\lambda=1.25$ as the final result since it has the best F1 Score.}
\label{lambda_ablation}
\vspace{-0.5cm}
\end{table}

\subsection{Dataset Generation}

\textbf{Object 3D Model Datasets.} The 3D models used for the simulation experiments come from three datasets: Stanford 3D Scanning Repository~\cite{krishnamurthy1996fitting}, Linemod~\cite{hinterstoisser2012model}, and HomebrewedDB~\cite{kaskman2019homebreweddb}. All objects are divided into 40 training objects (identical to those in~\cite{pan2023one}) and 10 test objects for experimentation as listed in Fig.~\ref{evp_metrics}(a). From the data collected from the 40 training objects, $80\%$ is allocated for network training, while the remaining 20$\%$ is used for validation.

\textbf{OSVP Dataset Generation.} The object sizes range from 0.05\,m to 0.15\,m. The view space size $n$ is set to 32 and the radius is set to 0.4\,m. The hyper-parameter $\alpha$ for set covering optimization is set to 10. The resolution of the point cloud acquired by virtual imaging of each view is set to 2\,mm, that is on the ground truth surface points and is considered as positive input to POCO. In addition, POCO requires negative inputs for training, which is sampled in free space around the surfaces. The parameters for POCO training are FKAConv backbone with latent size set to view space size $32$ and InterpAttentionKHeadsNet~\cite{Boulch_2022_CVPR} as the decoder. In total, we generate 14,375 labeled training data.

\subsection{OSVP Network Training}

\textbf{Implementation Details.} Our OSVP network utilizes the PointCloud Transformer in the PoinTr network with its default parameters. The ViewState Transformer is a 4 layers transformer followed by a fully connected layer. The outputs from both transformers have a dimension of 1,024. The predict layer, implemented by MLP, takes a 1,024-dimensional as input and returns a prediction $\hat{V}_{pred}$.

\textbf{Ablation Study and Results.} The results presented in Table~\ref{lambda_ablation} indicate that with a $\lambda$ value exceeding 1, the network tends to prioritize a high recall at the cost of reduced precision. We chose $\lambda = 1.25$ because of the highest F1 score. Fig.~\ref{evp_metrics}(b) shows the precision, recall, and loss on the validation dataset during OSVP training with $\lambda = 1.25$. 


\subsection{Evaluation of Efficiency of View Planning}\label{secV.C}

\textbf{Metrics and Baselines.} The \textit{surface coverage} is used to evaluate the quality of the observed sparse points while the number of \textit{required views} and \textit{movement cost} is used to evaluate the reconstruction efficiency. To make a broad comparison, three search-based NBV methods~\cite{delmerico2018comparison,pan2021global,pan2023global}, two learning-based NBV methods~\cite{mendoza2020supervised,zeng2020pc}, and two one-shot methods~\cite{pan2022scvp,pan2023one} are reported as baselines.

\textbf{Ablation Study and Results.} In order to determine the pipeline modules of our method, we perform an ablation study on three strategies as shown in Table~\ref{SC_ab}. According to these results, we adopt the approach with POCO refinement for next-stage comparison because of its highest surface coverage. Table~\ref{SC_nbv_oneshot} reports the final reconstruction results on observed accumulated surfaces. The results indicate that our method can achieve a reasonable surface coverage with fewer views and less movement cost compared to baselines. In terms of resource efficiency, our method outperforms baselines, achieving a higher surface coverage/resource ratio.

\textbf{Planning time.} For each test object, search-based methods~\cite{delmerico2018comparison,pan2021global,pan2023global} take around 20-30 seconds while learning-based methods~\cite{mendoza2020supervised,zeng2020pc,pan2022scvp,pan2023one} take less than 1 second. Our OSVP network takes less than 1 second but requires around 5-10 seconds for POCO refinement.

\begin{table}[!t]
\centering
\resizebox{0.90\columnwidth}{!}{%
\begin{tabular}{|c|c|c|c|}
\hline
Method & \begin{tabular}[c]{@{}c@{}}Surface\\Coverage\,(\%)\end{tabular}\,$\uparrow$ & \begin{tabular}[c]{@{}c@{}}Required\\Views\end{tabular}\,$\downarrow$ & \begin{tabular}[c]{@{}c@{}}Movement\\Costs\,(m)\end{tabular}\,$\downarrow$ \\ \hline
Ours-Sparse & 87.47\,±\,9.60 & \textbf{5.30}\,±\,1.57 & \textbf{1.46}\,±\,0.40 \\ \hline
Ours-NBV & 89.76\,±\,4.40 & 6.00\,±\,0.67 & 2.02\,±\,0.25 \\ \hline
Ours\,(Proposed) & \textbf{90.00}\,±\,4.57 & 5.80\,±\,1.03 & 1.59\,±\,0.19 \\ \hline
\end{tabular}
}
\caption{Ablation study on our pipeline modules. Ours-Sparse stands for initial sparse point cloud input into the OSVP network instead of Ours using POCO refined point cloud. Ours-NBV stands for adding one more NBV to observe more surfaces before POCO refinement, following the pipeline in MA-SCVP~\cite{pan2023one}. Each value reports the average mean and standard deviation on 10 test objects. As can be seen, Ours using POCO refinement improves the surface coverage and achieves less standard deviation compared to Ours-Sparse while adding NBV shows no benefit for our OSVP network.
}
\label{SC_ab}
\vspace{-0.2cm}
\end{table}

\begin{table*}[!t]
\centering
\resizebox{0.8\textwidth}{!}{%
\begin{tabular}{|c|c|c|c|c|c|}
\hline
Method & \begin{tabular}[c]{@{}c@{}}Surface\\Coverage\,(\%)\end{tabular}\,$\uparrow$ & \begin{tabular}[c]{@{}c@{}}Required\\Views\end{tabular}\,$\downarrow$ & \begin{tabular}[c]{@{}c@{}}Movement\\Costs\,(m)\end{tabular}\,$\downarrow$ & $\mathrm{\displaystyle\frac{Surface\ Coverage}{Required\ Views}}\,\uparrow$ & $\mathrm{\displaystyle\frac{Surface\ Coverage}{Movement\ Cost}}\,\uparrow$ \\ \hline
RSE~\cite{delmerico2018comparison} & 97.16\,±\,1.92 & 10 & 4.31\,±\,0.39 & 9.72\,±\,0.19 & 22.70\,±\,2.15 \\ \hline
MCMF~\cite{pan2021global} & 97.93\,±\,1.17 & 10 & 4.75\,±\,0.61 & 9.79\,±\,0.12 & 20.95\,±\,2.99 \\ \hline
GMC~\cite{pan2023global} & \textbf{98.20}\,±\,1.25 & 10 & 4.73\,±\,0.61 & 9.82\,±\,0.13 & 21.10\,±\,2.97 \\ \hline
NBVNet~\cite{mendoza2020supervised} & 93.98\,±\,9.19 & 10 & 5.15\,±\,0.50 & 9.40\,±\,0.92 & 18.39\,±\,2.47 \\ \hline
PCNBV~\cite{zeng2020pc} & 96.58\,±\,3.87 & 10 & 4.97\,±\,0.49 & 9.66\,±\,0.39 & 19.61\,±\,2.00 \\ \hline
MA-SCVP~\cite{pan2023one} & 96.59\,±\,4.07 & 12.40\,±\,1.26 & 3.32\,±\,0.23 & 7.85\,±\,0.70 & 29.25\,±\,2.43 \\ \hline
SCVP~\cite{pan2022scvp} & 88.69\,±\,13.89 & 9.60\,±\,0.70 & 2.15\,±\,0.09 & 9.24\,±\,1.33 & 41.38\,±\,6.73 \\ \hline
Ours & 90.00\,±\,4.57 & \textbf{5.80}\,±\,1.03 & \textbf{1.59}\,±\,0.19 & \textbf{15.98}\,±\,3.08 & \textbf{57.46}\,±\,7.23 \\ \hline
\end{tabular}
}
\caption{
Comparison of view planning reconstruction results on observed accumulated surfaces. We report the surface coverage per required views and per movement cost to evaluate the resource efficiency in terms of information gain. Each value reports the average mean and standard deviation on 10 test objects. Note that NBV-based methods run with default iterations, \ie fixed required views of 10, to achieve high surface coverage. As can be seen, our method achieves significantly higher resource efficiency compared to baselines.
}
\label{SC_nbv_oneshot}
\vspace{-0.2cm}
\end{table*}

\begin{table*}[!t]
\centering
\resizebox{0.8\textwidth}{!}{%
\begin{tabular}{|c|c|c|c|c|c|}
\hline
Method & \begin{tabular}[c]{@{}c@{}}Surface\\Coverage\,(\%)\end{tabular}\,$\uparrow$ & \begin{tabular}[c]{@{}c@{}}Movement\\Cost\,(m)\end{tabular}\,$\downarrow$ & \begin{tabular}[c]{@{}c@{}}Chamfer\\Distance\,(mm)\end{tabular}\,$\downarrow$ & \multicolumn{1}{c|}{\begin{tabular}[c]{@{}c@{}}Earth Mover's\\ Distance\,(mm)\end{tabular}\,$\downarrow$} & \multicolumn{1}{c|}{\begin{tabular}[c]{@{}c@{}}Density-aware\\Chamfer Distance\end{tabular}\,$\downarrow$} \\ \hline
GMC~\cite{pan2023global} & 88.70\,±\,12.61 & 2.55\,±\,0.52 & 4.078\,±\,1.604 & 4.514\,±\,1.954 & 0.4745\,±\,0.0376 \\ \hline
PCNBV~\cite{zeng2020pc} & \textbf{92.47}\,±\,4.97 & 2.87\,±\,0.59 & 3.888\,±\,1.207 & 3.937\,±\,1.204 & 0.4715\,±\,0.0408 \\ \hline
MA-SCVP~\cite{pan2023one} & 82.47\,±\,9.93 & 1.62\,±\,0.33 & 4.383\,±\,1.282 & 4.373\,±\,1.418 & 0.4844\,±\,0.0423 \\ \hline
SCVP~\cite{pan2022scvp} & 75.77\,±\,14.41 & \textbf{1.23}\,±\,0.29 & 4.731\,±\,1.658 & 4.869\,±\,1.967 & 0.4964\,±\,0.0471 \\ \hline
Ours & 90.00\,±\,4.57 & 1.59\,±\,0.19 & \textbf{3.885}\,±\,1.273 & \textbf{3.895}\,±\,1.141 & \textbf{0.4662}\,±\,0.0274 \\ \hline
\end{tabular}
}
\caption{
Comparison of view planning results on mesh reconstruction under a limited view budget. Each method finishes the reconstruction under the limited number of views as our OSVP network outputs (5.80\,±\,1.03). Each value reports the average mean and standard deviation on 10 test objects. Note that MA-SCVP and SCVP are not designed in a way to stop early, showing a low surface coverage with less movement cost. As can be seen, our method achieves better mesh metrics even with lower surface coverage compared to PCNBV, indicating that we observe a sparse point cloud suitable for better implicit reconstruction.
}
\label{VB}
\vspace{-0.5cm}
\end{table*}

\begin{table}[!t]
\centering
\resizebox{0.95\columnwidth}{!}{%
\begin{tabular}{|c|c|c|c|}
\hline
Method & \begin{tabular}[c]{@{}c@{}}Chamfer\\Distance\,(mm)\end{tabular}\,$\downarrow$ & \multicolumn{1}{c|}{\begin{tabular}[c]{@{}c@{}}Earth Mover's\\Distance\,(mm)\end{tabular}\,$\downarrow$} & \multicolumn{1}{c|}{\begin{tabular}[c]{@{}c@{}}Density-aware\\Chamfer Distance\end{tabular}\,$\downarrow$} \\ \hline
Poisson~\cite{kazhdan2006poisson} & 7.142\,±\,3.292 & 15.889\,±\,5.053 & 0.6113\,±\,0.0402 \\ \hline
BPA~\cite{bernardini1999ball} & 3.899\,±\,1.189 & 7.933\,±\,2.244 & 0.5215\,±\,0.0461 \\ \hline
POCO~\cite{Boulch_2022_CVPR} & \textbf{3.885}\,±\,1.273 & \textbf{3.895}\,±\,1.141 & \textbf{0.4662}\,±\,0.0274 \\ \hline
\end{tabular}
}
\caption{Ablation study on mesh reconstruction methods. Ball-Pivoting Algorithm~(BPA)~\cite{bernardini1999ball} is an explicit method while Poisson~\cite{kazhdan2006poisson} is an implicit method. As an implicit neural representation, POCO can effectively improve the mesh quality, confirming the ability to obtain small missing surfaces without observing them.
}
\label{ab_MR}
\vspace{-0.2cm}
\end{table}

\begin{figure}[!t]
\centering
\includegraphics[width=1.0\columnwidth]{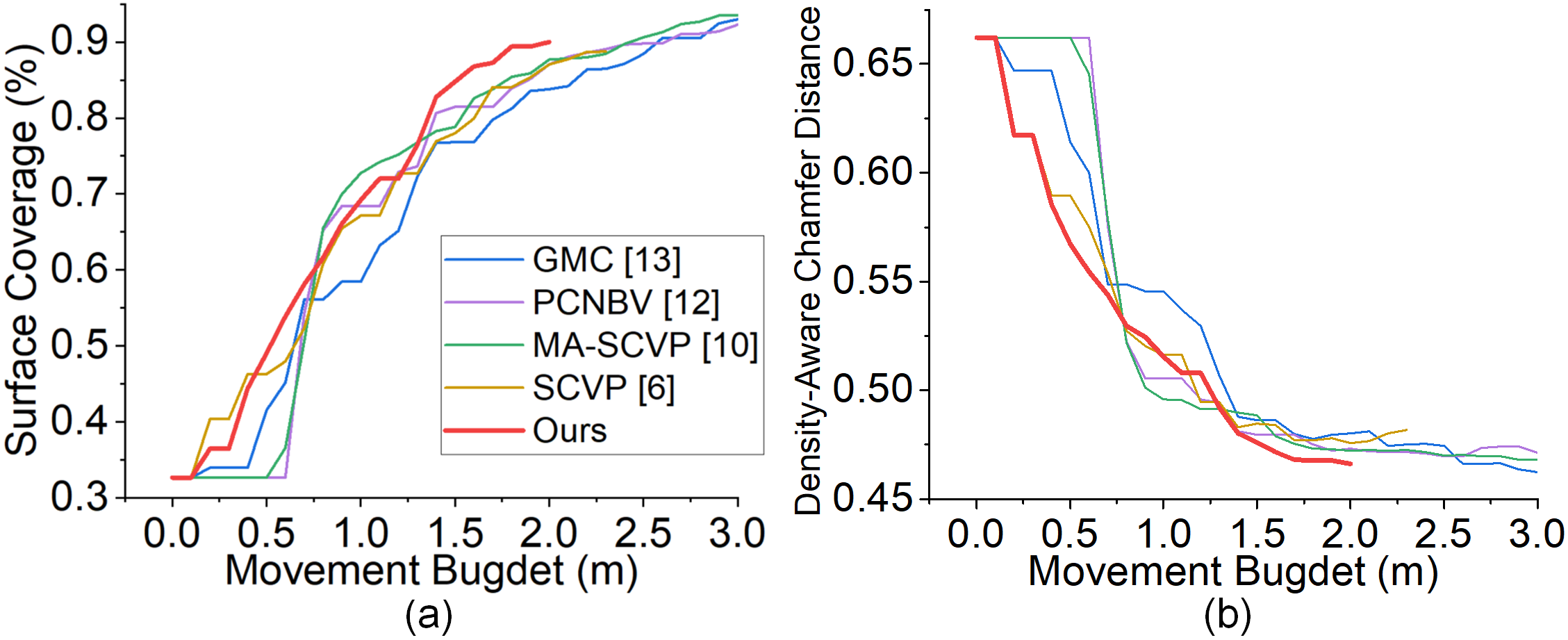}
\caption{Comparison of view planning results on mesh reconstruction under limited movement budgets. The curves are drawn with an interval movement budget value of 0.1\,m. For a certain movement budget value, each method is allowed to plan views until its movement cost exceeds this movement budget to report the reconstruction results. Note that some methods finish their reconstruction early, making their curve short. As can be seen, our method achieves higher surface coverage and lower density-aware Chamfer distance within a small movement budget compared to other baselines.
}
\label{fig_limited_movement}
\vspace{-0.5cm}
\end{figure}

\subsection{Mesh Reconstruction under Limited Resource}

\textbf{Evaluation of Implicit Surface.} To evaluate the quality of implicit surfaces and to compare it with other explicit methods, we follow the evaluation in POCO~\cite{Boulch_2022_CVPR}, which performs comparison on mesh rendered from implicit surfaces trained with sensor-visible empty points~\cite{sulzer2022deep}.

\textbf{Metrics.} Since we have the ground truth mesh for reference, three common mesh distance metrics~\cite{wu2021densityaware} are considered: Chamfer Distance~(CD), Earth Mover's Distance~(EMD), and Density-aware Chamfer Distance~(DCD). They are used to measure the similarity between rendered mesh and ground truth mesh, focusing on structural and global similarity. Note that DCD solves some limitations in CD and EMD, avoiding local mismatch insensitivity and neglecting detailed structure fidelity. All metrics are calculated on 10,240 sampled points on meshes.

\begin{figure}[!t]
\centering
\includegraphics[width=1.0\columnwidth]{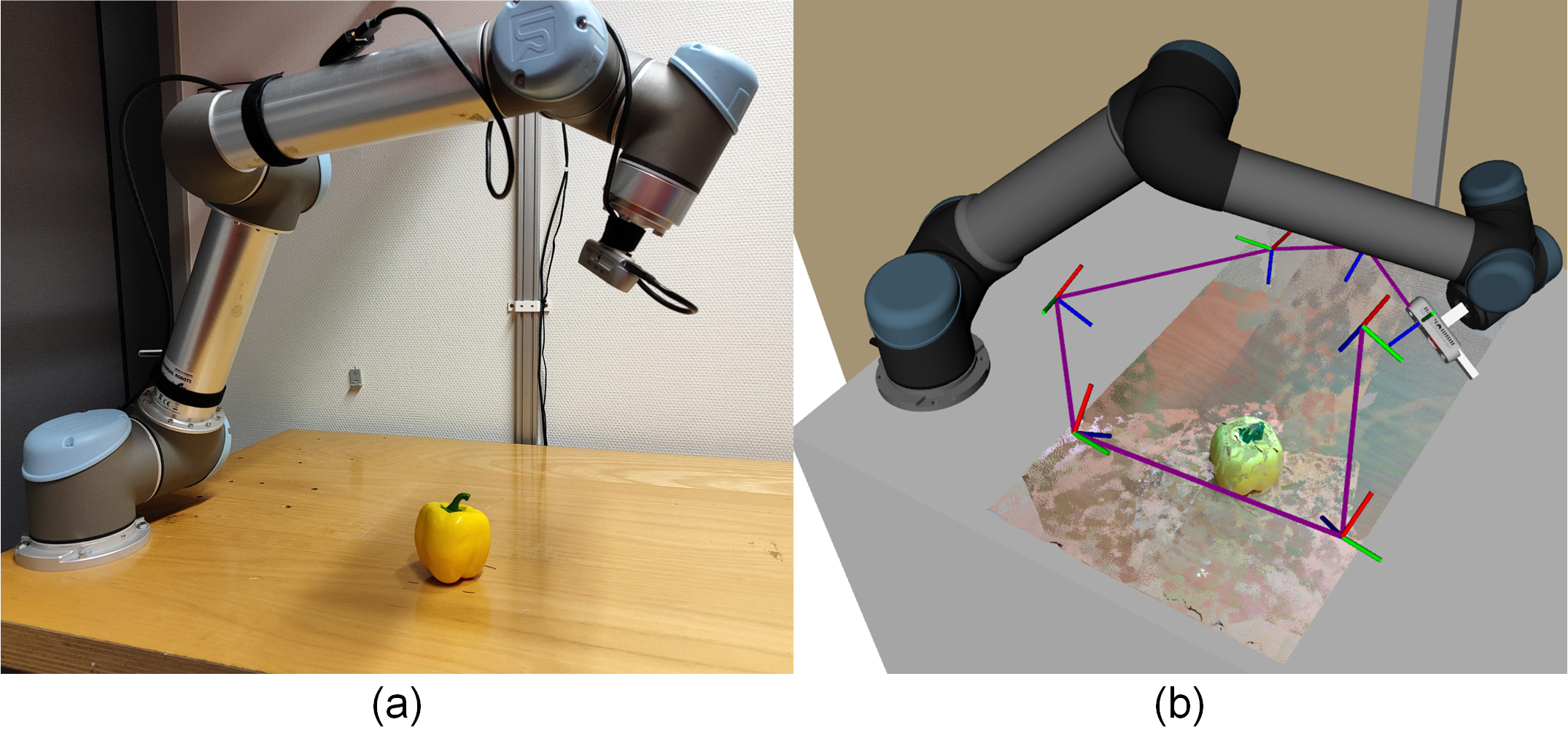}
\caption{Real-world reconstruction: (a) robot environment and the object, (b) planned global path (purple) and accumulated observed point clouds. As can be seen, our OSVP network predicts a small number of views (seven views) to cover most surface areas of an unknown object with low movement cost.}
\label{fig_realworld}
\vspace{-0.5cm}
\end{figure}

\textbf{Ablation Study and Results.} There are two types of resources usually considered in a robotic reconstruction task: the number of required views and movement cost. Comparative results with a limited view budget are shown in Table~\ref{VB}. Our method exhibits better mesh metrics even with a lower surface coverage compared to PCNBV, indicating that our sparse observed point cloud is better suited for implicit reconstruction. Fig.~\ref{fig_limited_movement} shows comparative results under a limited movement budget. Our method achieves the most efficient view planning regarding both surface and mesh reconstruction metrics. In situations with limited resources, our information gain is particularly higher. Finally, to analyze the impact of mesh reconstruction methods, an ablation study is reported in Table~\ref{ab_MR}. Compared to two traditional mesh reconstruction methods, POCO shows superior performance as an interpolation method.

\subsection{Real-World Experiment}

We deploy our approach in a real-world environment using a UR5 robot arm with an Intel Realsense D435 camera mounted on its end-effector and use ROS~\cite{koubaa2017robot} with MoveIt~\cite{chitta2016moveit} for robotic motion planning. We conduct a reconstruction test and demonstrate the ability to obtain a sparse but sufficient point cloud for POCO meshing with a small number of views. The reconstruction process and results are shown in Fig.~\ref{fig_realworld} and in the accompanying video at \url{https://youtu.be/h24cc2wYAoc}.

\section{Conclusion}
\label{sec:conclusion}

In this paper, we present an efficient view planning method for active object reconstruction, taking advantage of both the movement efficiency of one-shot pipelines and the ability to obtain small missing surfaces by a convolutional implicit representation. We propose using refined dense point clouds for set covering instead of sparse observation point clouds to adapt the one-shot pipeline to POCO. Our OSVP network is trained on our new dataset to predict a small set of views. We plan a globally connected path among the set of views of the network output representing the minimum movement cost. Simulated experiments show that by using the OSVP network, our method achieves a better reconstruction quality compared to several state-of-the-art baselines under limited view and movement budgets. The real-world experiment suggests the applicability of our method.

\bibliographystyle{IEEEtranN}
\footnotesize
\bibliography{bibliography}

\begin{thebibliography}{45}
\providecommand{\natexlab}[1]{#1}
\providecommand{\url}[1]{#1}
\csname url@samestyle\endcsname
\providecommand{\newblock}{\relax}
\providecommand{\bibinfo}[2]{#2}
\providecommand{\BIBentrySTDinterwordspacing}{\spaceskip=0pt\relax}
\providecommand{\BIBentryALTinterwordstretchfactor}{4}
\providecommand{\BIBentryALTinterwordspacing}{\spaceskip=\fontdimen2\font plus
\BIBentryALTinterwordstretchfactor\fontdimen3\font minus
  \fontdimen4\font\relax}
\providecommand{\BIBforeignlanguage}[2]{{%
\expandafter\ifx\csname l@#1\endcsname\relax
\typeout{** WARNING: IEEEtranN.bst: No hyphenation pattern has been}%
\typeout{** loaded for the language `#1'. Using the pattern for}%
\typeout{** the default language instead.}%
\else
\language=\csname l@#1\endcsname
\fi
#2}}
\providecommand{\BIBdecl}{\relax}
\BIBdecl

\bibitem[Chen et~al.(2011)Chen, Li, and Kwok]{chen2011active}
S.~Chen, Y.~Li, and N.~M. Kwok, ``Active vision in robotic systems: A survey of
  recent developments,'' \emph{Intl.~Journal~of Robotics Research (IJRR)},
  vol.~30, no.~11, pp. 1343--1377, 2011.

\bibitem[Song et~al.(2022)Song, Kim, and Choi]{song2021view}
S.~Song, D.~Kim, and S.~Choi, ``{View Path Planning via Online Multiview Stereo
  for 3-D Modeling of Large-Scale Structures},'' \emph{IEEE Trans.~on Robotics
  (TRO)}, vol.~38, pp. 372 --390, 2022.

\bibitem[Zeng et~al.(2020{\natexlab{a}})Zeng, Wen, Zhao, and Liu]{zeng2020view}
R.~Zeng, Y.~Wen, W.~Zhao, and Y.-J. Liu, ``View planning in robot active
  vision: A survey of systems, algorithms, and applications,''
  \emph{Computational Visual Media}, pp. 1--21, 2020.

\bibitem[Delmerico et~al.(2018)Delmerico, Isler, Sabzevari, and
  Scaramuzza]{delmerico2018comparison}
J.~Delmerico, S.~Isler, R.~Sabzevari, and D.~Scaramuzza, ``{A comparison of
  volumetric information gain metrics for active 3D object reconstruction},''
  \emph{Autonomous Robots}, vol.~42, no.~2, pp. 197--208, 2018.

\bibitem[Pan and Wei(2021)]{pan2021global}
S.~Pan and H.~Wei, ``{A Global Max-Flow-Based Multi-Resolution Next-Best-View
  Method for Reconstruction of 3D Unknown Objects},'' \emph{IEEE Robotics and
  Automation Letters (RA-L)}, vol.~7, no.~2, pp. 714--721, 2021.

\bibitem[Pan et~al.(2022)Pan, Hu, and Wei]{pan2022scvp}
S.~Pan, H.~Hu, and H.~Wei, ``Scvp: Learning one-shot view planning via set
  covering for unknown object reconstruction,'' \emph{IEEE Robotics and
  Automation Letters}, vol.~7, no.~2, pp. 1463--1470, 2022.

\bibitem[Kaba et~al.(2017)Kaba, Uzunbas, and Lim]{kaba2017reinforcement}
M.~D. Kaba, M.~G. Uzunbas, and S.~N. Lim, ``{A Reinforcement Learning Approach
  to the View Planning Problem},'' in \emph{Proc.~of the IEEE Conf.~on Computer
  Vision and Pattern Recognition (CVPR)}, 2017.

\bibitem[Pan and Wei(2022)]{pan2022aglobal}
S.~Pan and H.~Wei, ``A global max-flow-based multi-resolution next-best-view
  method for reconstruction of 3d unknown objects,'' \emph{IEEE Robotics and
  Automation Letters}, vol.~7, no.~2, pp. 714--721, 2022.

\bibitem[Boulch and Marlet(2022)]{Boulch_2022_CVPR}
A.~Boulch and R.~Marlet, ``{POCO: Point Convolution for Surface
  Reconstruction},'' in \emph{Proc.~of the IEEE Conf.~on Computer Vision and
  Pattern Recognition (CVPR)}, 2022.

\bibitem[Pan et~al.(2023)Pan, Hu, Wei, Dengler, Zaenker, and
  Bennewitz]{pan2023one}
S.~Pan, H.~Hu, H.~Wei, N.~Dengler, T.~Zaenker, and M.~Bennewitz, ``One-shot
  view planning for fast and complete unknown object reconstruction,''
  \emph{arXiv preprint arXiv:2304.00910}, 2023.

\bibitem[Mendoza et~al.(2020)Mendoza, Vasquez-Gomez, Taud, Sucar, and
  Reta]{mendoza2020supervised}
M.~Mendoza, J.~I. Vasquez-Gomez, H.~Taud, L.~E. Sucar, and C.~Reta,
  ``Supervised learning of the next-best-view for 3d object reconstruction,''
  \emph{Pattern Recognition Letters}, vol. 133, pp. 224--231, 2020.

\bibitem[Zeng et~al.(2020{\natexlab{b}})Zeng, Zhao, and Liu]{zeng2020pc}
R.~Zeng, W.~Zhao, and Y.-J. Liu, ``{PC-NBV: A Point Cloud Based Deep Network
  for Efficient Next Best View Planning},'' in \emph{Proc.~of the IEEE/RSJ
  Intl.~Conf.~on Intelligent Robots and Systems (IROS)}, 2020.

\bibitem[{}Pan and Wei(2023)]{pan2023global}
S.~{}Pan and H.~Wei, ``A global generalized maximum coverage-based solution to
  the non-model-based view planning problem for object reconstruction,''
  \emph{Computer Vision and Image Understanding}, vol. 226, p. 103585, 2023.

\bibitem[Peng et~al.(2020)Peng, Niemeyer, Mescheder, Pollefeys, and
  Geiger]{peng2020convolutional}
S.~Peng, M.~Niemeyer, L.~Mescheder, M.~Pollefeys, and A.~Geiger,
  ``Convolutional occupancy networks,'' in \emph{Proc.~of the Europ.~Conf.~on
  Computer Vision (ECCV)}, 2020.

\bibitem[Tang et~al.(2021)Tang, Lei, Xu, Ma, Jia, and Zhang]{tang2021sa}
J.~Tang, J.~Lei, D.~Xu, F.~Ma, K.~Jia, and L.~Zhang, ``{SA-ConvONet:
  Sign-Agnostic Optimization of Convolutional Occupancy Networks},'' in
  \emph{Proc.~of the IEEE Intl.~Conf.~on Computer Vision (ICCV)}, 2021.

\bibitem[Potthast and Sukhatme(2014)]{potthast2014probabilistic}
C.~Potthast and G.~S. Sukhatme, ``A probabilistic framework for next best view
  estimation in a cluttered environment,'' \emph{Journal of Visual
  Communication and Image Representation}, vol.~25, no.~1, pp. 148--164, 2014.

\bibitem[Daudelin and Campbell(2017)]{daudelin2017adaptable}
J.~Daudelin and M.~Campbell, ``An adaptable, probabilistic, next-best view
  algorithm for reconstruction of unknown 3-d objects,'' \emph{IEEE Robotics
  and Automation Letters (RA-L)}, vol.~2, no.~3, pp. 1540--1547, 2017.

\bibitem[Zaenker et~al.(2021)Zaenker, Smitt, McCool, and
  Bennewitz]{zaenker2021viewpoint}
T.~Zaenker, C.~Smitt, C.~McCool, and M.~Bennewitz, ``Viewpoint planning for
  fruit size and position estimation,'' in \emph{Proc.~of the IEEE/RSJ
  Intl.~Conf.~on Intelligent Robots and Systems (IROS)}, 2021.

\bibitem[Menon et~al.(2023)Menon, Zaenker, and Bennewitz]{menon2022viewpoint}
R.~Menon, T.~Zaenker, and M.~Bennewitz, ``{NBV-SC: Next Best View Planning
  based on Shape Completion for Fruit Mapping and Reconstruction},'' in
  \emph{Proc.~of the IEEE/RSJ Intl.~Conf.~on Intelligent Robots and Systems
  (IROS)}, 2023.

\bibitem[Zaenker et~al.(2023)Zaenker, R{\"u}ckin, Menon, Popovi{\'c}, and
  Bennewitz]{zaenker2023graph}
T.~Zaenker, J.~R{\"u}ckin, R.~Menon, M.~Popovi{\'c}, and M.~Bennewitz,
  ``Graph-based view motion planning for fruit detection,'' in \emph{Proc.~of
  the IEEE/RSJ Intl.~Conf.~on Intelligent Robots and Systems (IROS)}, 2023.

\bibitem[Peralta et~al.(2020)Peralta, Casimiro, Nilles, Aguilar, Atienza, and
  Cajote]{peralta2020next}
D.~Peralta, J.~Casimiro, A.~M. Nilles, J.~A. Aguilar, R.~Atienza, and
  R.~Cajote, ``{Next-best view policy for 3D reconstruction},'' in
  \emph{Proc.~of the Europ.~Conf.~on Computer Vision (ECCV)}, 2020.

\bibitem[Zeng et~al.(2022)Zeng, Zaenker, and Bennewitz]{zeng2022deep}
X.~Zeng, T.~Zaenker, and M.~Bennewitz, ``{Deep Reinforcement Learning for
  Next-Best-View Planning in Agricultural Applications},'' in \emph{Proc.~of
  the IEEE Intl.~Conf.~on Robotics \& Automation (ICRA)}, 2022.

\bibitem[Dengler et~al.(2023)Dengler, Pan, Kalagaturu, Menon, Dawood, and
  Bennewitz]{dengler2023viewpoint}
N.~Dengler, S.~Pan, V.~Kalagaturu, R.~Menon, M.~Dawood, and M.~Bennewitz,
  ``Viewpoint push planning for mapping of unknown confined spaces,'' in
  \emph{Proc.~of the IEEE/RSJ Intl.~Conf.~on Intelligent Robots and Systems
  (IROS)}, 2023.

\bibitem[Wu et~al.(2014)Wu, Sun, Long, Huang, Cohen-Or, Gong, Deussen, and
  Chen]{wu2014quality}
S.~Wu, W.~Sun, P.~Long, H.~Huang, D.~Cohen-Or, M.~Gong, O.~Deussen, and
  B.~Chen, ``Quality-driven poisson-guided autoscanning,'' \emph{ACM
  Transactions on Graphics}, vol.~33, no.~6, 2014.

\bibitem[Mildenhall et~al.(2020)Mildenhall, Srinivasan, Tancik, Barron,
  Ramamoorthi, and Ng]{mildenhall2021nerf}
B.~Mildenhall, P.~P. Srinivasan, M.~Tancik, J.~T. Barron, R.~Ramamoorthi, and
  R.~Ng, ``{NeRF: Representing Scenes as Neural Radiance Fields for View
  Synthesis},'' \emph{Proc.~of the Europ.~Conf.~on Computer Vision (ECCV)},
  2020.

\bibitem[Park et~al.(2019)Park, Florence, Straub, Newcombe, and
  Lovegrove]{park2019deepsdf}
J.~J. Park, P.~Florence, J.~Straub, R.~Newcombe, and S.~Lovegrove, ``{DeepSDF:
  Learning Continuous Signed Distance Functions for Shape Representation},'' in
  \emph{Proc.~of the IEEE Conf.~on Computer Vision and Pattern Recognition
  (CVPR)}, 2019, pp. 165--174.

\bibitem[Yan et~al.(2023)Yan, Liu, Quan, Chen, and Fu]{yan2023active}
D.~Yan, J.~Liu, F.~Quan, H.~Chen, and M.~Fu, ``{Active Implicit Object
  Reconstruction using Uncertainty-guided Next-Best-View Optimziation},''
  \emph{arXiv preprint arXiv:2303.16739}, 2023.

\bibitem[S{\"u}nderhauf et~al.(2023)S{\"u}nderhauf, Abou-Chakra, and
  Miller]{sunderhauf2023density}
N.~S{\"u}nderhauf, J.~Abou-Chakra, and D.~Miller, ``Density-aware nerf
  ensembles: Quantifying predictive uncertainty in neural radiance fields,'' in
  \emph{Proc.~of the IEEE Intl.~Conf.~on Robotics \& Automation (ICRA)}, 2023.

\bibitem[Ran et~al.(2023)Ran, Zeng, He, Chen, Li, Chen, Lee, and
  Ye]{ran2023neurar}
Y.~Ran, J.~Zeng, S.~He, J.~Chen, L.~Li, Y.~Chen, G.~Lee, and Q.~Ye, ``{NeurAR:
  Neural Uncertainty for Autonomous 3D Reconstruction With Implicit Neural
  Representations},'' \emph{IEEE Robotics and Automation Letters (RA-L)},
  vol.~8, no.~2, pp. 1125--1132, 2023.

\bibitem[Lai et~al.(2023)Lai, Yue, Hao, Glover, and L{\"u}]{lai2023iterated}
X.~Lai, D.~Yue, J.-K. Hao, F.~Glover, and Z.~L{\"u}, ``Iterated dynamic
  neighborhood search for packing equal circles on a sphere,'' \emph{Computers
  \& Operations Research}, vol. 151, p. 106121, 2023.

\bibitem[Held and Karp(1962)]{held1962dynamic}
M.~Held and R.~M. Karp, ``A dynamic programming approach to sequencing
  problems,'' \emph{Journal of the Society for Industrial and Applied
  mathematics}, vol.~10, no.~1, pp. 196--210, 1962.

\bibitem[Chang et~al.(2015)Chang, Funkhouser, Guibas, Hanrahan, Huang, Li,
  Savarese, Savva, Song, Su, Xiao, Yi, and Yu]{shapenet2015}
A.~X. Chang, T.~Funkhouser, L.~Guibas, P.~Hanrahan, Q.~Huang, Z.~Li,
  S.~Savarese, M.~Savva, S.~Song, H.~Su, J.~Xiao, L.~Yi, and F.~Yu,
  ``{ShapeNet: An Information-Rich 3D Model Repository},'' \emph{arXiv preprint
  arXiv:1512.03012}, 2015.

\bibitem[Hornung et~al.(2013)Hornung, Wurm, Bennewitz, Stachniss, and
  Burgard]{hornung2013octomap}
A.~Hornung, K.~M. Wurm, M.~Bennewitz, C.~Stachniss, and W.~Burgard, ``{OctoMap:
  An efficient probabilistic 3D mapping framework based on octrees},''
  \emph{Autonomous Robots}, vol.~34, no.~3, pp. 189--206, 2013.

\bibitem[Gurobi~Optimization(2021)]{gurobi2021gurobi}
L.~Gurobi~Optimization, ``Gurobi optimizer reference manual,'' 2021.

\bibitem[Yu et~al.(2021)Yu, Rao, Wang, Liu, Lu, and Zhou]{yu2021pointr}
X.~Yu, Y.~Rao, Z.~Wang, Z.~Liu, J.~Lu, and J.~Zhou, ``{PoinTr: Diverse Point
  Cloud Completion with Geometry-Aware Transformers},'' in \emph{Proc.~of the
  IEEE Intl.~Conf.~on Computer Vision (ICCV)}, 2021.

\bibitem[Vaswani et~al.(2017)Vaswani, Shazeer, Parmar, Uszkoreit, Jones, Gomez,
  Kaiser, and Polosukhin]{vaswani2017attention}
A.~Vaswani, N.~Shazeer, N.~Parmar, J.~Uszkoreit, L.~Jones, A.~N. Gomez,
  {\L}.~Kaiser, and I.~Polosukhin, ``Attention is all you need,''
  \emph{Advances in neural information processing systems}, vol.~30, 2017.

\bibitem[Krishnamurthy and Levoy(1996)]{krishnamurthy1996fitting}
V.~Krishnamurthy and M.~Levoy, ``Fitting smooth surfaces to dense polygon
  meshes,'' in \emph{Proc.~ of the 23rd annual conference on Computer graphics
  and interactive techniques}, 1996, pp. 313--324.

\bibitem[Hinterstoisser et~al.(2012)Hinterstoisser, Lepetit, Ilic, Holzer,
  Bradski, Konolige, and Navab]{hinterstoisser2012model}
S.~Hinterstoisser, V.~Lepetit, S.~Ilic, S.~Holzer, G.~Bradski, K.~Konolige, and
  N.~Navab, ``Model based training, detection and pose estimation of
  texture-less 3d objects in heavily cluttered scenes,'' in \emph{accv}, 2012.

\bibitem[Kaskman et~al.(2019)Kaskman, Zakharov, Shugurov, and
  Ilic]{kaskman2019homebreweddb}
R.~Kaskman, S.~Zakharov, I.~Shugurov, and S.~Ilic, ``{HomebrewedDB: RGB-D
  Dataset for 6D Pose Estimation of 3D Objects},'' in \emph{Proc.~of the IEEE
  Conf.~on Computer Vision and Pattern Recognition (CVPR)}, 2019.

\bibitem[Kazhdan et~al.(2006)Kazhdan, Bolitho, and Hoppe]{kazhdan2006poisson}
M.~Kazhdan, M.~Bolitho, and H.~Hoppe, ``Poisson surface reconstruction,'' in
  \emph{Proc.~of the Eurographics Symposium on Geometry Processing}, vol.~7,
  2006.

\bibitem[Bernardini et~al.(1999)Bernardini, Mittleman, Rushmeier, Silva, and
  Taubin]{bernardini1999ball}
F.~Bernardini, J.~Mittleman, H.~Rushmeier, C.~Silva, and G.~Taubin, ``The
  ball-pivoting algorithm for surface reconstruction,'' \emph{IEEE Trans.~on
  Visualization and Computer Graphics}, vol.~5, no.~4, pp. 349--359, 1999.

\bibitem[Sulzer et~al.(2022)Sulzer, Landrieu, Boulch, Marlet, and
  Vallet]{sulzer2022deep}
R.~Sulzer, L.~Landrieu, A.~Boulch, R.~Marlet, and B.~Vallet, ``Deep surface
  reconstruction from point clouds with visibility information,'' in \emph{2022
  26th International Conference on Pattern Recognition (ICPR)}.\hskip 1em plus
  0.5em minus 0.4em\relax IEEE, 2022, pp. 2415--2422.

\bibitem[Wu et~al.(2021)Wu, Pan, Zhang, WANG, Liu, and Lin]{wu2021densityaware}
T.~Wu, L.~Pan, J.~Zhang, T.~WANG, Z.~Liu, and D.~Lin, ``Density-aware chamfer
  distance as a comprehensive metric for point cloud completion,'' in \emph{In
  Advances in Neural Information Processing Systems (NeurIPS)}, 2021.

\bibitem[Koub{\^a}a et~al.(2017)]{koubaa2017robot}
A.~Koub{\^a}a \emph{et~al.}, \emph{Robot Operating System (ROS).}\hskip 1em
  plus 0.5em minus 0.4em\relax Springer, 2017.

\bibitem[Chitta(2016)]{chitta2016moveit}
S.~Chitta, ``Moveit!: an introduction,'' \emph{Robot Operating System (ROS) The
  Complete Reference (Volume 1)}, pp. 3--27, 2016.

\end{thebibliography}

\end{document}